\documentclass[10pt]{article}
\usepackage[a4paper, margin=1.2in]{geometry}
\usepackage{amsmath, amssymb}
\usepackage{graphicx}
\usepackage{float}
\usepackage{hyperref}
\usepackage{amsmath}
\usepackage{amsfonts}
\usepackage{amssymb}
\usepackage{subcaption}
\usepackage{algorithm}
\usepackage{algpseudocode}
\usepackage{array}
\usepackage{xcolor}
\usepackage{cite}
\usepackage[utf8]{inputenc}
\usepackage{booktabs} 
\usepackage{array}    
\usepackage{multirow,multicol} 
\usepackage{caption}  
\usepackage{float}   
\date{\vspace{-5ex}} 

\title{Intelligent ROI-Based Vehicle Counting Framework for Automated Traffic Monitoring}
\author{Mohamed A. Abdelwahab$^{1,4}$, Zaynab Al-Ariny$^{2,3}$, Mahmoud Fakhry$^{3,5}$, El-Sayed Hasaneen$^{3}$\\[6pt]
\small $^{1}$Electrical Engineering Dept., Faculty of Energy Engineering, Aswan University, Aswan, Egypt\\
\small $^{2}$Electrical Engineering Dept., Faculty of Engineering, Sohag University, Sohag, Egypt\\
\small $^{3}$Electrical Engineering Dept., Faculty of Engineering, Aswan University, Aswan, Egypt\\
\small $^{4}$Information Technology Dept., New Assiut Technological University (NATU), Assiut, Egypt\\
    \small $^{5}$CEIEC Research Institute, Universidad Francisco de Vitoria, Madrid, Spain\\
\small \texttt{abdelwahab@aswu.edu.eg}, \texttt{zenab\_khalf@eng.sohag.edu.eg}, \texttt{mahmoud.fakhry@ufv.es}
, \texttt{hasaneen@aswu.edu.eg}}

\begin{document}

\maketitle

\begin{abstract}
Accurate vehicle counting through video surveillance is crucial for efficient traffic management. However, achieving high counting accuracy while ensuring computational efficiency remains a challenge. To address this, we propose a fully automated, video-based vehicle counting framework designed to optimize both computational efficiency and counting accuracy. Our framework operates in two distinct phases: \textit{estimation} and \textit{prediction}. In the estimation phase, the optimal region of interest (ROI) is automatically determined using a novel combination of three models based on detection scores, tracking scores, and vehicle density. This adaptive approach ensures compatibility with any detection and tracking method, enhancing the framework's versatility. In the prediction phase, vehicle counting is efficiently performed within the estimated ROI. We evaluated our framework on benchmark datasets like UA-DETRAC, GRAM, CDnet 2014, and ATON. Results demonstrate exceptional accuracy, with most videos achieving 100\% accuracy, while also enhancing computational efficiency—making processing up to four times faster than full-frame processing. The framework outperforms existing techniques, especially in complex multi-road scenarios, demonstrating robustness and superior accuracy. These advancements make it a promising solution for real-time traffic monitoring.
\end{abstract}


\section{Introduction}
Traffic congestion continues to challenge urban environments, impacting commuters, city planners, and environmental sustainability. As cities evolve into smart ecosystems, Traffic Management Systems (TMS) have become essential for reducing congestion, optimizing travel times, and improving urban quality of life. Central to TMS is the ability to monitor traffic flow, plan infrastructure, and support data-driven decisions. Vehicle counting, a key component of TMS, goes beyond quantifying vehicles—it is about rethinking urban mobility and creating safer, more livable cities.

Recent advancements in deep learning have enabled highly accurate vehicle counting methods \cite{santos2020, gomaa2022, narayanan2025, youssef2021}. However, these methods often process entire video frames, leading to high computational overhead and increased processing time. To address this, researchers have explored restricting detection or tracking regions to reduce complexity and improve efficiency. For example, some studies narrowed the counting region to mitigate the identity-switch problem \cite{huang2023}, where trackers incorrectly assign new IDs to vehicles, improving counting accuracy. Shuang et al. \cite{li2020} proposed counting vehicles using a horizontal line near the frame's center, limiting tracking to the nearest n vehicles. Similarly, \cite{li2021} defined the ROI as the area closest to the camera, where vehicles appear larger, and used a perpendicular counting line. Other approaches \cite{abdelwahab2019, alariny2020} relied on manually selecting the ROI, while M. Abbas et al. \cite{abbas2020} introduced a GUI for defining movements of interest (MOI). Lijun et al. \cite{yu2020} further simplified the process by using a predefined ROI.

Despite their high accuracy and efficiency, these methods \cite{huang2023, li2020, li2021, abdelwahab2019, alariny2020, abbas2020, yu2020} share a critical limitation: reliance on manually selected counting lines or ROIs. Such manual interventions lack universal applicability, as the optimal counting area varies across scenarios. For instance, the best counting region is not always near the camera or the frame's center, as previously assumed. This necessitates manual adaptation for different scenarios, reducing flexibility and scalability.

To address these limitations, some studies have proposed automated methods for ROI selection. Malolan et al. \cite{vasu2021} introduced an automated vehicle counting method that dynamically determines the ROI based on the highest detection confidence. However, the method exhibited significant performance variability across videos, indicating limited generalization capabilities. Hadi et al. \cite{ghahremannezhad2020} developed a motion-based approach to automatically define the ROI as the entire road area where moving vehicles are detected. While innovative, this approach does not focus on a specific smaller region, potentially increasing computational load. Yang et al. \cite{he2022} proposed an automatic ROI selection method based on filtering pixels with high texture complexity. However, this method also lacks generalizability, as the optimal counting region is not always the most congested area.

In summary, while existing vehicle counting methods often achieve high accuracy, they face significant challenges, including high computational demands, reliance on manual adjustments, and limited adaptability across diverse environments. These limitations hinder their practicality for real-time applications and underscore the need for a more robust, automated, and computationally efficient solution.
\begin{figure}[t!]
    \centering
    \includegraphics[width=0.58\linewidth]{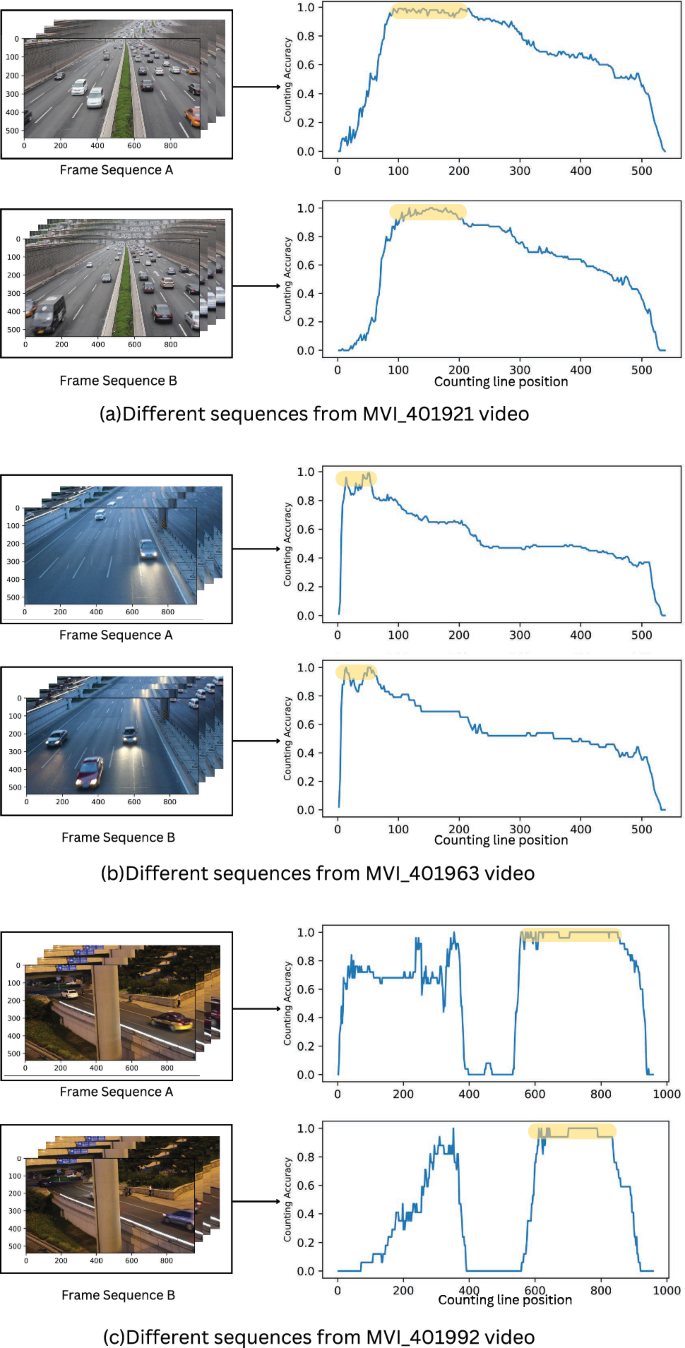}
    \caption{The optimal region for vehicle counting varies considerably between videos. Interestingly, each video demonstrates strong internal consistency, with the best region-highlighted in yellow-remaining relatively stable throughout the video. (a), (b) and (c) show the counting accuracy of different videos from the UA-DETRAC dataset}
    \label{fig:fig1}
\end{figure}

\subsection{Paper Motivation \& Contributions}
Selecting an optimal region of interest (ROI) for vehicle counting that ensures consistent accuracy in various scenarios is a challenge. Fig. \ref{fig:fig1} illustrates accuracy curves for three video scenarios, obtained by placing multiple counting lines perpendicular to traffic flow. The curves reveal significant accuracy variations between videos, indicating that no universal optimal ROI exists. However, within individual videos, accuracy curves for different sequences (A and B) show similar patterns, suggesting that for a fixed scene we can effectively determine an ROI for accurate counting. This paper proposes a novel vehicle counting framework based on automatic ROI selection, prioritizing three key characteristics:

\begin{enumerate}
    \item \textbf{Automation:} The framework operates autonomously across diverse scenes, eliminating the need for manual adjustments and ensuring consistent performance in various environments.
    \item \textbf{Efficiency:} By focusing on a limited region of interest (ROI), the framework minimizes computational power and processing time, making it highly suitable for real-time applications.
    \item \textbf{Generality:} The framework seamlessly adapts to any camera scene without requiring customization. It employs techniques that automatically extract relevant parameters from the video feed, ensuring broad applicability.
\end{enumerate}

By emphasizing these characteristics, our framework overcomes the limitations of existing methods, providing an efficient, automated, and adaptable solution for real-time vehicle counting. It dynamically identifies the optimal ROI using three critical metrics derived automatically from detection and tracking outputs: detection score, tracking score, and detection density. Then, counting is performed within the selected ROI, ensuring both adaptability and computational efficiency.

The remainder of this paper is organized as follows: Section 2 provides a detailed description of the proposed framework. Section 3 presents and analyzes the experimental results. Finally, Section 4 concludes the paper and discusses potential future directions.

\begin{figure}[t!]
    \centering
    \includegraphics[width=0.6\linewidth]{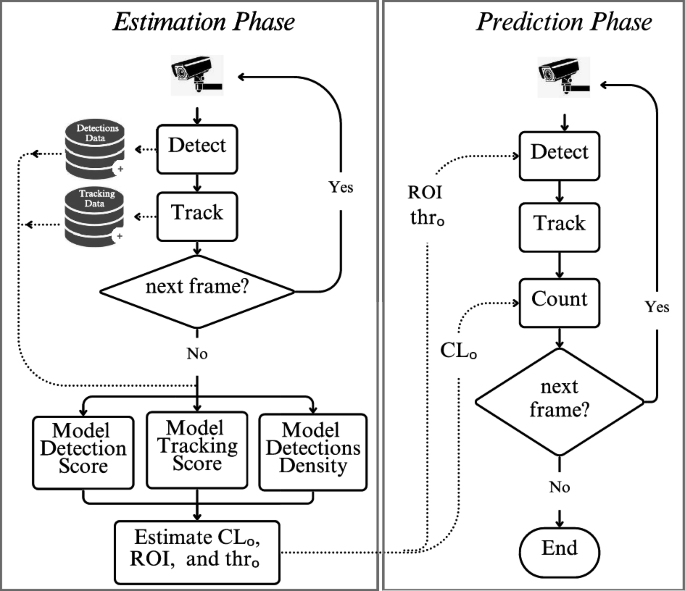}
    \caption{Key components of the proposed AIR-VC framework, including
estimation and prediction phases}
    \label{fig:fig2}
\end{figure}

\section{Proposed AIR-VC Framework}
In this paper, we introduce the \textbf{Automated, Intelligent, ROI-based Vehicle Counting (AIR-VC)} framework. AIR-VC is designed to intelligently and automatically determine the optimal region of interest (ROI) for efficient vehicle counting. The framework is highly versatile, supporting integration with any detection and tracking methods. For this study, we implement \textit{YOLOv8} \cite{jocher2023} for detection and \textit{DeepSORT} \cite{wojke2017} for tracking, owing to their demonstrated high accuracy as validated in \cite{sohan2024}.

The AIR-VC framework operates in two key phases: \textit{estimation} and \textit{prediction}. During the estimation phase, a segment of the input frame sequence is used to develop three models based on detection scores, tracking scores, and vehicle density. These models are employed to estimate critical parameters that identify the optimal ROI for counting. In the subsequent prediction phase, the framework performs vehicle counting using the estimated parameters. Fig. \ref{fig:fig2} provides an overview of the key steps in the proposed AIR-VC framework.

\subsection{Estimation Phase}
In this phase, we analyze tracking and detection scores from a frame sequence segment to estimate the optimal counting line, ROI, and detection confidence threshold. Implementation involves the following steps:

\subsubsection{Detection Score Modeling}
A high detection score (detection confidence) is critical for achieving accurate vehicle counting. However, detection performance is influenced by various factors, including video resolution, weather conditions, camera placement, and illumination. Therefore, the optimal region of interest (ROI) should correspond to the area where the detection score is maximized.

To model the detection score, detection is performed on every frame within the estimation sequence for all vehicles. For each detected vehicle, the bounding box (BBox) is identified, and the following information is recorded: the detection score, the BBox center coordinates ($cx$, $cy$), the BBox height, and the BBox width. Once all detections are completed, the direction of the counting lines ($CLs$) is determined to be perpendicular to the direction of vehicle movement. This direction is derived by analyzing the differences in the center coordinates ($cx$, $cy$) of detected BBoxes across consecutive frames.

To ensure the robustness of the model, outlier detection scores are removed using the Interquartile Range (IQR) method \cite{porter2005}. This step prevents outliers from adversely affecting the average detection score and the accuracy of the regression model \cite{osborne2004, blatna2006}. Subsequently, polynomial regression \cite{ostertagova2012} is employed to statistically model the relationship between the average detection score $D_s$ (as the dependent variable) and the $CL$ position $p$ (as the independent variable). The $CL$ position refers to position at X/Y axis for horizontal/vertical traffic flow. This relationship is expressed as a polynomial equation of order $n$ as follows:

\begin{equation}
D_s = a_0 + a_1 p + a_2 p^2 + \ldots + a_n p^n
\label{eq:detection_score}
\end{equation}

Where: $ D_s $ is the average detection score, $ p $ is the $ CL $ position, and $ a_0, a_1, \ldots, a_n $ are the coefficients of the polynomial equation. This model aims to identify the area with the highest detection score, which is most likely to contain the optimal counting line (CL). To evaluate the performance of the regression model, the Root Mean Square Error (RMSE) is utilized, as it is a widely recognized and valuable metric for assessing model accuracy \cite{bera2021, khan2022}. Following the suggestion of Singh et al. \cite{singh2005}, RMSE values less than half the standard deviation of the measured data (i.e., the detection score values) are considered sufficiently low, indicating a well-fitted model. Using this approach, we select the most appropriate polynomial order $ n $ to balance model complexity and accuracy, ensuring an optimal trade-off between overfitting and underfitting.
\begin{figure}[t!]
    \centering
    \includegraphics[width=0.8\linewidth]{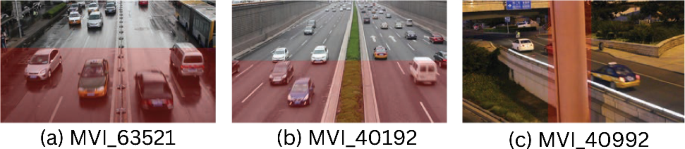}
    \caption{The camera-limited regions (highlighted in red) for different scenarios from the UA-DETRAC dataset. For the area near the camera, the camera angle restricts the visibility of the leftmost and rightmost lanes in (a) and (b). In (c), an obstacle hides the middle area of the frame}
    \label{fig:fig3}
\end{figure}
\subsubsection{Tracking Score Modeling}
While a high detection score ensures accurate identification of vehicles, a high tracking score is essential for correctly tracking these vehicles across consecutive frames. This prevents double-counting and maintains object identities, which are critical for achieving accurate vehicle counting. Multi-object tracking (MOT) \cite{jimenez2022} is a technique used to identify and track the movement of multiple objects in a video sequence. In this study, we employ DeepSORT \cite{wojke2017}, which utilizes Mahalanobis distance for motion information and cosine distance for feature information. These distances are represented as cost matrices, with tracking scores calculated as $ 1 - Cost $. The final tracking score is computed as the sum of the motion score and the feature score.

Following the same steps outlined in Section 2.1.1, the average tracking score is calculated for each counting line (CL). The relationship between the average tracking score $ T_s $ and the CL position $ p $ is modeled using a polynomial equation of order $ m $, as shown in Eq. (\ref{eq:tracking_score}). Here, $ T_s $ represents the average tracking score, $ p $ is the CL position, $ m $ is the order of the polynomial equation, and $ a_0, a_1, \ldots, a_m $ are the coefficients of the polynomial equation.

\begin{equation}
T_s = a_0 + a_1 p + a_2 p^2 + \ldots + a_m p^m
\label{eq:tracking_score}
\end{equation}

\subsubsection{Detections Density Modeling}
While detection and tracking scores help exclude regions with blurry scenes or heavily overlapping vehicles, areas that typically yield low scores, relying solely on these metrics to determine the optimal counting line ($ CL $) is insufficient. In various video scenarios, camera-limited regions arise due to obstacles or camera angles that restrict the visibility of vehicles in certain parts of the frame. For instance, as illustrated by the red-highlighted areas in Fig. \ref{fig:fig3}, the camera angle in MVL\_63521 and MVL\_40192 misses both the leftmost and rightmost lanes in regions near the camera. Similarly, in MVL\_40992, an obstacle blocks the middle area of the scene. Selecting the $ CL $ within such camera-limited regions would make it impossible to count all vehicles, even if the detection and tracking scores are high.

To address this issue, we introduce a new metric: \textbf{Detections Density}. This metric quantifies the number of vehicles detected across the entire frame. By analyzing detections density, we can identify the \textbf{Highest Detections Density Region (HDDR)}, which represents the area where the largest number of vehicles are detected throughout the frame. Selecting the optimal counting line ($ cl_o $) within this region ensures that it is not located in a camera-limited area, thereby significantly improving counting accuracy.

\begin{figure}[t!]
    \centering
    \includegraphics[width=0.6\linewidth]{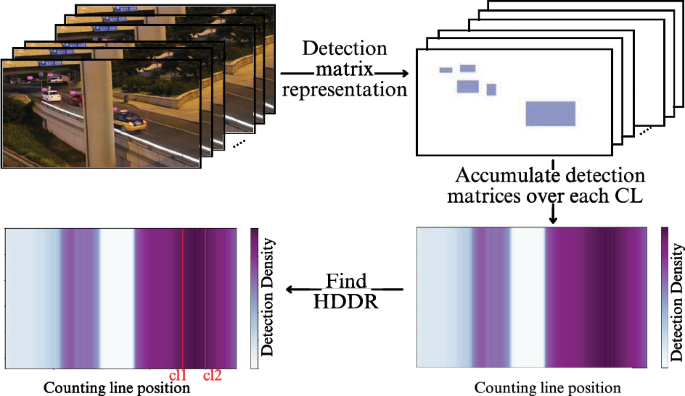}
    \caption{Identifying the HDDR: A Heatmap is generated using the previously obtained detection scores. The HDDR (region within $cl_1$ and $cl_2$ ) is identified according to the generated heat map. The HDDR is
represented by the wider region with the most vehicle detections}
    \label{fig:fig4}
\end{figure}

To model detections density, we utilize the detection scores obtained from Section 2.1.1 to identify the HDDR. This is achieved by representing each detected vehicle as a matrix of ones with the same dimensions as its bounding box (BBox). By summing these detection matrices over each counting line ($ CL $) across video frames, we generate a heat map. As illustrated in Fig. \ref{fig:fig4}, the heat map highlights regions with high detections density (indicated by darker colors), revealing areas where the majority of vehicles are likely to pass through the frame. The HDDR is then identified as the widest region with the highest intensity, defined by the boundaries ($ cl_1, cl_2 $).

\subsubsection{Optimal CL Finding}
\begin{figure}[t!]
    \centering
    \includegraphics[width=0.7\linewidth]{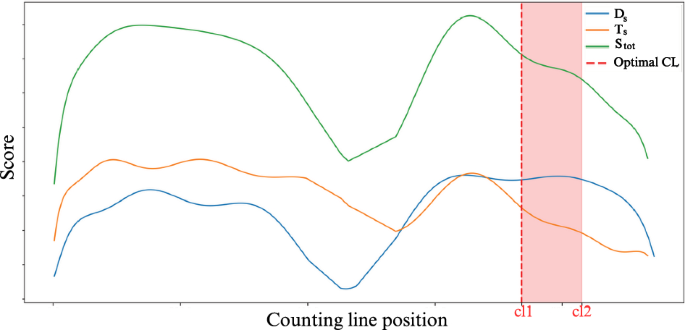}
    \caption{Selecting the optimal counting line(the red dashed line) within the HDDR which indicated by ($cl_1$ ,$cl_2$)}
    \label{fig:fig5}
\end{figure}

As shown in Fig. \ref{fig:fig5}, we seek the $ cl_o $ by maximizing the value of the overall score $ S_{tot} $, expressed in Eq. (\ref{eq:total_score}), within the HDDR. Using Brute Force Algorithm \cite{heule2017}, we exhaustively explore all possibilities within the range ($ cl_1, cl_2 $) to find $ cl_o $ with the maximum value of $ S_{tot} $. Consequently, we select the optimal counting line at $ cl_o $.

\begin{equation}
S_{tot} = D_s + T_s
\label{eq:total_score}
\end{equation}

\begin{figure}[t!]
    \centering
    \includegraphics[width=0.7\linewidth]{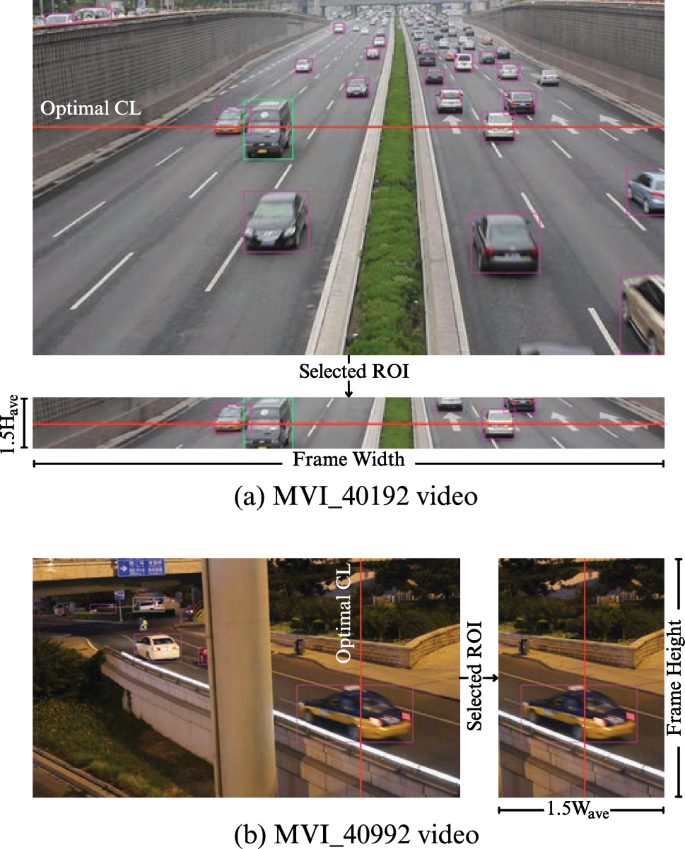}
    \caption{Samples of selected ROI for videos from UA-DETRAC Dataset. Samples show videos with (a) vertical and (b) horizontal traffic flow}
    \label{fig:fig6}
\end{figure}
\subsubsection{ROI Selection}
Building on Section 2.1.4, we define a region of interest (ROI) centered around the optimal counting line ($ cl_o $) for detection and tracking tasks, significantly reducing processing time and computational load compared to full-frame analysis. The ROI is dynamically determined based on detected vehicle bounding box (BBox) dimensions at $ cl_o $. As shown in Fig. \ref{fig:fig6}, for vertical traffic flow, the ROI height is $ \alpha $ times the average BBox height ($ H_{ave} $), with width matching the frame width. For horizontal flow, the ROI height equals the frame height, and its width is $ \alpha $ times the average BBox width ($ W_{ave} $). Using $ \alpha = 1.5 $, a 0.5-times margin ensures sufficient coverage for accurate detection while minimizing unnecessary processing.

\subsubsection{$Thr_o$ Selecting}
The detection confidence score reflects the model's certainty in correctly identifying an object within an image. The detection confidence threshold ($ thr_o $) defines the minimum score a prediction must achieve to be considered valid; predictions below this threshold are discarded. The choice of $ thr_o $ critically impacts detection performance \cite{wenkel2021}, as it directly influences the trade-off between false positives (incorrect detections) and false negatives (undetected vehicles).

In this step, the optimal $ thr_o $ is determined using detection scores collected in Sec 2.1.1. By analyzing scores within the ROI boundaries and filtering out outliers, the lowest score is identified as $ thr_o $ and applied for all subsequent detections.

\subsection{Prediction Phase}
In this phase, the estimated parameters ($ cl_o $, ROI, and $ thr_o $), determined during the Estimation Phase, are used to automatically predict the vehicle count. Vehicle detection is restricted to the selected ROI, and $ thr_o $ is applied to filter out vehicles with insufficient detection confidence. Each detected vehicle within the ROI is tracked using an MOT algorithm and assigned a unique ID. As each tracked vehicle crosses the designated $ cl_o $, its ID is recorded, increasing the vehicle count. By restricting the counting process to the ROI, the risk of identity-switch \cite{huang2023} is effectively eliminated, ensuring reliable and accurate counting.

\section{Results}
In our experiments, YOLOv8 \cite{jocher2023} was utilized for detection within the cropped ROI without resizing, and a confidence threshold of 0.2 was applied during the estimation phase. To evaluate the proposed AIR-VC framework, we conducted experiments on a diverse video scenarios.

\subsection{Datasets}
Different video datasets were used to evaluate AIR-VC. The M-30 sequence from the GRAM dataset \cite{guerrero2013} was recorded on a sunny day at 800×480 resolution and 30 fps. The Highway sequence from CDnet 2014 \cite{wang2014} (320×240) includes strong vehicle and tree shadows. HighwayII from ATON \cite{trivedi2000} shares similar characteristics but also features minor camera jitter and a crowded scene. Four sequences from UA-DETRAC \cite{wen2020}(MVL\_63521, MVL\_40992, MVL\_40963, MVL\_40192) were captured at (960×540) resolution and 25 fps. MVL\_63521 presents rainy conditions and heavy occlusion, MVL\_40992 features night-time illumination, MVL\_40963 has strong headlights in a night view, and MVL\_40192 depicts heavy occlusion under cloudy conditions. Each video sequence was split into approximately 70\% for estimation and 30\% for prediction.
\begin{table}[t!]
\centering
\caption{Processing speed (FPS) using the entire frame and the proposed AIR-VC for each video sequence. $h \times w$ refers to the sequence resolution (Height $\times$ Width). The last column shows the speed improvement achieved by using AIR-VC compared to using the entire frame.}
\label{tab:table1}
\begin{tabular}{lccccccc}
\hline
Sequence & \multicolumn{2}{l}{Entire Frame} & &\multicolumn{2}{l}{AIR-VC} & \multicolumn{2}{c}{Speed } \\
\cline{2-3} \cline{5-6}
         & $h \times w$ & FPS & &$h \times w$ & FPS & Imp. \\
\hline
M\_30        & 480 $\times$ 800& 32  && 98 $\times$ 800& 114   & 255\% \\
Highway      & 240 $\times$ 320& 26  && 112 $\times$ 320& 53   & 103\% \\
HighwayII    & 240 $\times$ 320& 26  && 36 $\times$ 320& 115   & 337\% \\
MVI\_63521   & 540 $\times$ 960& 32  && 99 $\times$ 960& 114   & 258\% \\
MVI\_40992   & 540 $\times$ 960& 26  && 540 $\times$ 376& 32   & 23\% \\
MVI\_40963   & 540 $\times$ 960& 32  && 94 $\times$ 960& 162   & 403\% \\
MVI\_40192   & 540 $\times$ 960& 32  && 78 $\times$ 960& 162   & 406\% \\
\hline
\end{tabular}
\end{table}
\subsection{Processing Speed Evaluation}
In this experiment, we evaluated the AIR-VC framework's processing speed, measured in Frames Per Second (FPS). The analysis compared AIR-VC to methods processing the entire frame. The speed improvement is calculated using Eq. (\ref{eq:speed_improvement}), where $ Speed\, Imp. $ is the speed improvement, $ FPS_{ROI} $ is the FPS using AIR-VC, and $ FPS_{frame} $ is the FPS using the entire frame:

\begin{equation}
Speed\, Imp. = \frac{FPS_{ROI} - FPS_{frame}}{FPS_{frame}} \times 100\%
\label{eq:speed_improvement}
\end{equation}

As shown in Table \ref{tab:table1}, AIR-VC achieves a significant speed improvement of up to 400\% compared to full-frame processing.
\begin{figure}[t!]
    \centering
    \includegraphics[width=0.6\linewidth]{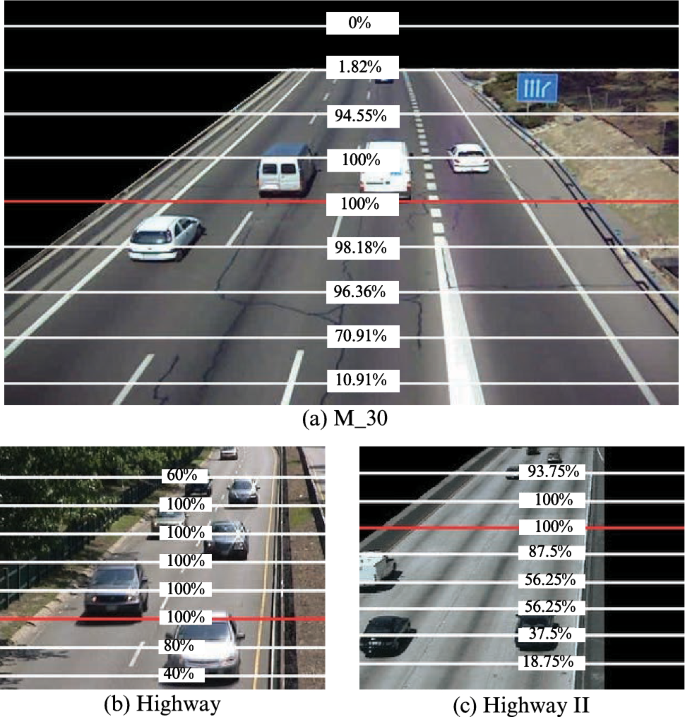}
    \caption{Counting accuracy for M-30, Highway and HighwayII video sequences using $cl_o$ selected by the proposed AIR-VC (red line) and other uniformly distributed counting lines across the frame (white lines)}
    \label{fig:fig7}
\end{figure}
\begin{figure}[t!]
    \centering
    \includegraphics[width=1\linewidth]{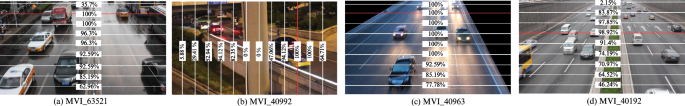}
    \caption{Counting accuracy for video sequences from UA-DETRAC dataset using clo selected by the proposed AIR-VC (red line) and other uniformly distributed counting lines across the frame (white lines)}
    \label{fig:fig8}
\end{figure}
\subsection{Counting Accuracy Evaluation}
In this experiment, we compared the counting accuracy using $ cl_o $ estimated by AIR-VC with that of uniformly distributed lines across the frame. As shown in Figs. \ref{fig:fig7} and \ref{fig:fig8}, $ cl_o $ consistently achieves the highest count accuracy in all video sequences. The significant variation in accuracy across frame regions highlights the challenge of defining a universally optimal counting area, emphasizing the need for an automated ROI selection technique like the proposed AIR-VC framework.

\subsection{Comparison with State-of-the-Art Methods}
By comparing the counting accuracy of the proposed method with state-of-the-art methods on the same video sequences, our approach demonstrates excellent performance. As shown in Table \ref{tab:table2}, AIR-VC achieves perfect accuracy (100\%) for the M-30, Highway, and HighwayII videos. Results in Table \ref{tab:table3} further reveal that AIR-VC achieves the highest accuracy across all UA-DETRAC dataset videos, except for MVL\_40192, where its accuracy is slightly lower (by only 0.18\%) than the best result reported by \cite{li2021}. Despite this minor difference, AIR-VC remains highly competitive. While some methods achieve comparable accuracy for specific videos, AIR-VC stands out by offering fully automated processing and superior efficiency.

\begin{table}[ht]
\centering
\caption{Counting accuracy comparison for M-30, Highway and HighwayII video sequences.}
\label{tab:table2}
\begin{tabular}{lcccc}
\hline
Sequence & M-30 & Highway & HighwayII \\
lanes    & 4    & 2       & 4 \\
\hline

S. Li~\cite{li2020}        & 100\% & 96.3\% & N/A \\
Abdelwahab~\cite{abdelwahab2019} & 93.51\% & 96.43\% & 95.83\% \\
Zaynab~\cite{alariny2020}  & 98.7\% & 100\% & 97.9\% \\
A. Gomaa~\cite{gomaa2022}  & 100\% & 100\% & 97.9\% \\
Adson~\cite{santos2020}    & 98.7\% & 100\% & 97.92\% \\
Proposed AIR-VC & \textbf{100\%} & \textbf{100\%} & \textbf{100\%} \\
\hline
\end{tabular}
\end{table}
\begin{table}[ht]
\centering
\caption{Counting accuracy comparison for video sequences from UA-DETRAC dataset.}
\label{tab:table3}
\begin{tabular}{lcccc}
\hline
\multicolumn{2}{l}{Sequence} &S. Li~\cite{li2021} & Yang~\cite{yang2021} & Proposed \\
\cline{1-2}
name & lanes &  & &AIR-VC \\
\hline
MVI\_63521 & 3+3 & 97.56\% & N/A & \textbf{100\%} \\
MVI\_40992 & 2   & N/A     & N/A & \textbf{100\%} \\
MVI\_40963 & 5   & 98.57\% & N/A & \textbf{100\%} \\
MVI\_40192 & 4+4 & 99.08\% & 97.96\% & 98.9\% \\
\hline
\end{tabular}
\end{table}
\begin{table}[t!]
\centering
\caption{Counting accuracy for M-30, HighwayII and MVI\_40963 video sequences considering multi-roads scenario using the proposed AIR-VC.}
\label{tab:table4}
\begin{tabular}{lccc}
\hline
\multicolumn{2}{l}{Sequence} && Accuracy \\
\cline{1-2}
name & lanes &&  \\
\hline
M-30        & 3+3+4 && 99.1\% \\
HighwayII   & 3+4   && 85.7\% \\
MVI\_40963  & 1+5+3 && 98.1\% \\
\hline
\end{tabular}
\end{table}
\begin{figure}[t!]
    \centering
    \includegraphics[width=0.64\linewidth]{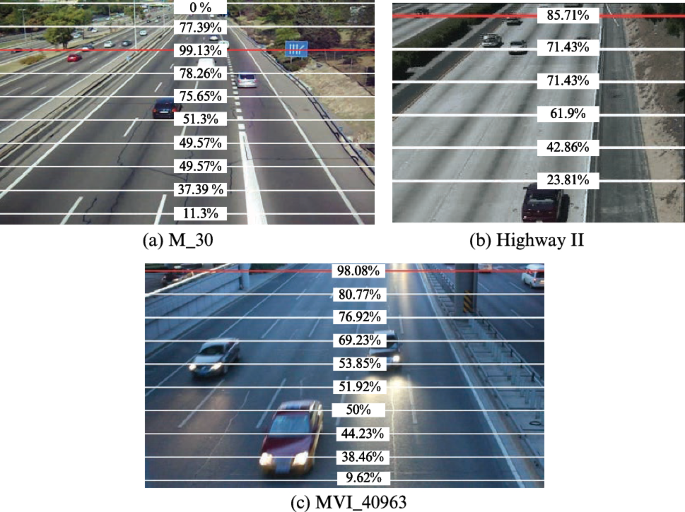}
    \caption{Counting accuracy considering multi-roads scenario using $cl_o$ selected by the proposed AIR-VC (red line) and other uniformly distributed counting lines across the frame (white lines)}
    \label{fig:fig9}
\end{figure}
\subsection{Multi-Roads Scenarios}
Previous studies \cite{santos2020, gomaa2022, li2020, li2021, abdelwahab2019, alariny2020} often focus on counting vehicles on a single main road, ignoring other visible side roads within the frame. To further validate AIR-VC's effectiveness and applicability, we applied it to the M-30, HighwayII, and MVL\_0963 video sequences, considering all visible roads and counting vehicles across all lanes—10 lanes for M-30, 7 for HighwayII, and 9 for MVL\_0963. As shown in Fig. \ref{fig:fig9}, AIR-VC successfully selected the optimal counting line and achieved the highest counting accuracy in each sequence. Notably, even for vehicles far from the camera, the automatically selected $ thr_o $ ensured acceptable counting accuracy. These results are summarized in Table \ref{tab:table4}.

\section{Conclusion}
In conclusion, this paper presents a fully automated and novel vehicle counting framework that effectively balances accuracy and computational efficiency. By leveraging three models, focused on detection scores, tracking scores, and vehicle density, we automate the process of identifying the optimal region of interest (ROI), counting line, and detection confidence threshold. This approach enables our framework to achieve high counting accuracy while significantly reducing computational costs, as demonstrated by the experimental results.

Our line-based counting framework is effectively evaluated using YOLOv8 for object detection and DeepSORT for tracking. However, the framework is designed to be flexible and can integrate with any other effective detection and tracking methods. The results highlight the framework's robustness and effectiveness across various video datasets, making it a valuable tool for modern traffic management systems.

To further enhance the proposed framework, we plan to extend it to handle more complex traffic scenarios, including intersections and crossroad layouts, which require multiple counting lines to accurately capture all passing vehicles. Applying such an extension, as the one suggested in \cite{rashid2025}, will involve developing algorithms capable of automatically identifying scenes with varying lane alignments, enabling the system to count vehicles separately in each direction. Such an enhancement would significantly broaden the framework's applicability and improve its accuracy in complex traffic environments.

\section*{Author Contributions}
Conceptualization: Mohamed Abdelwahab, Zaynab Al-Ariny; Methodology: Mohamed Abdelwahab, Zaynab Al-Ariny; Formal analysis and investigation: Mohamed Abdelwahab, Zaynab Al-Ariny; Writing - original draft preparation: Zaynab Al-Ariny; Writing - review and editing: Mohamed Abdelwahab, Mahmoud Fakhry, El-Sayed Hasaneen; Supervision: Mohamed Abdelwahab, Mahmoud Fakhry, El-Sayed Hasaneen.

\section*{Data Availability}
No datasets were generated or analysed during the current study.

\section*{Conflicts of interest}
The authors declare no competing interests.

\bibliographystyle{unsrt}
\bibliography{references}

\end{document}